\pdfoutput=1

\documentclass[11pt]{article}

\usepackage{EMNLP2023}

\usepackage{times}
\usepackage{latexsym}
\usepackage{graphicx}
\usepackage{tcolorbox}

\usepackage{tabularx}

\usepackage[T1]{fontenc}

\usepackage[utf8]{inputenc}

\usepackage{microtype}

\usepackage{inconsolata}

\title{Benchmarking the Legal Reasoning of LLMs in Arabic Islamic Inheritance Cases}

\author{Nouar AlDahoul \\
  Computer Science Department \\
   New York University \\
  Abu Dhabi, UAE \\
  \texttt{nouar.aldahoul@nyu.edu} \\\And
  Yasir Zaki \\
  Computer Science Department \\
   New York University \\
  Abu Dhabi, UAE \\
  \texttt{yasir.zaki@nyu.edu} \\}

\begin{document}
\maketitle
\begin{abstract}
Islamic inheritance domain holds significant importance for Muslims to ensure fair distribution of shares between heirs. Manual calculation of shares under numerous scenarios is complex, time-consuming, and error-prone. Recent advancements in Large Language Models (LLMs) have sparked interest in their potential to assist with complex legal reasoning tasks. This study evaluates the reasoning capabilities of state-of-the-art LLMs to interpret and apply Islamic inheritance laws. We utilized the dataset proposed in the ArabicNLP QIAS 2025 challenge, which includes inheritance case scenarios given in Arabic and derived from Islamic legal sources. Various  base and fine-tuned models, are assessed on their ability to accurately identify heirs, compute shares, and justify their reasoning in alignment with Islamic legal principles.
Our analysis reveals that the proposed majority voting solution, leveraging three base models (Gemini Flash 2.5, Gemini Pro 2.5, and GPT o3), outperforms all other models that we utilized across every difficulty level. It achieves up to 92.7\% accuracy and secures third place overall in the challenge~\cite{Task_1_Leaderboard_QIAS2025}.

\end{abstract}

\section{Introduction}

Islamic inheritance, which is known as ``Ilm al-Mawārīth'' in Arabic, is an area of jurisprudence that is highly structured, rule-based, and sensitive to context~\cite{abbasi2025ai_inheritance,qias2025}. The Qur'an introduced various rights and restrictions related to inheritance, marking significant improvements in the treatment of women and family relations for its time~\cite{IslamicInheritanceJurisprudence}. It also aimed to establish clear and fixed inheritance laws, contributing to the formation of a comprehensive legal system.

Islamic inheritance jurisprudence aims to prevent disputes by clearly defining the shares of each heir~\cite{islamicinheritance}. It ensures fair and equitable distribution, though Qur'anic verses assign different shares to specific relatives. Inheritance domain holds significant importance for Muslims, as it determines the rightful heirs, the individuals to be inherited from, and the specific shares allocated to each heir~\cite{zouaoui2021islamic}. Upon a person's death, a matter of particular concern is the management of all the property left behind. Manual calculation is a complex, time-consuming, and error-prone task that can be extremely difficult and costly. Automation of this calculation is convenient to save time, effort, and cost~\cite{zouaoui2021islamic}. 

Our analyses and experiments center around the
following research questions: \textbf{RQ1}: Do current Arabic open-source LLMs perform well in Islamic inheritance reasoning? \textbf{RQ2}: To what extent do state-of-the-art proprietary base LLMs excel in Islamic inheritance reasoning? and \textbf{RQ3}: Can fine-tuning LLMs for the inheritance reasoning task improve performance?

We address \textbf{RQ1} by running several open-source Arabic LLMs. To answer \textbf{RQ2}, we utilized APIs of state-of-the-art proprietary LLMs. Additionally, to answer \textbf{RQ3}, we fine-tuned several LLMs with the inheritance multiple-choice questions dataset.

\section{Related Work}

LLMs have shown impressive performance in a variety of natural language understanding tasks~\cite{aldahoul2024polytc,aldahoul2024advancing,kuo2025neutralizing}. When it comes to representing Islam, it is essential to ensure that its beliefs and teachings are portrayed accurately and faithfully, grounded in the Quran and Sunnah~\cite{patel2023building}. Additionally, it is important to prevent hallucination in Islamic fatwa generation~\cite{mohammed2025aftina}.

Several studies have focused on automating the inheritance calculation. \cite{jimoh2014design} used an expert system to calculate shares based on Islamic law. Despite the growing development of automated knowledge retrieval systems, few leverage semantic web technologies for Islamic knowledge, particularly in Arabic. \cite{zouaoui2021islamic} introduced AraFamOnto, an Arabic ontology-based system designed to automate Islamic inheritance calculations by efficiently modeling family relationships and reducing manual effort.

\cite{abbasi2025ai_inheritance} examined the capabilities of generative AI models, such as ChatGPT-4~\cite{GPT-4o}, Gemini~\cite{team2023gemini}, Co-Pilot~\cite{copilot2025}, and DeepSeek~\cite{deepseek2024}, in applying the principles of Islamic inheritance law. In their experiment, although ChatGPT-4 surpasses other models in performance, it continues to exhibit notable limitations, including fabricated references and legal inaccuracies. There is an ongoing effort to transform Islamic studies through generative AI as part of the 2024–2027 project~\cite{islamgpt2025}. This initiative focuses on building AI tools to engage with classical and contemporary Islamic texts.

However, previous efforts to automate inheritance problem solving using generative models remain limited, both in the number of models explored and due to the absence of large-scale datasets encompassing diverse scenarios and difficulty levels. This study addresses these gaps by utilizing a comprehensive dataset~\cite{QIAS2025_2,bouchekif2024inheritance} of inheritance cases to evaluate the performance of state-of-the-art LLMs.

\section{Materials and Methods}

\subsection{Dataset Overview}

\begin{figure}
    \centering
    \includegraphics[width=1\linewidth]{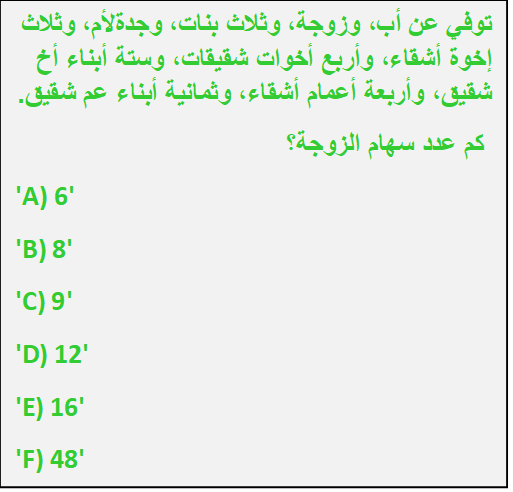}
    \caption{Example 1}
    \label{fig:exm1}
\end{figure}

\begin{figure}
    \centering
    \includegraphics[width=1\linewidth]{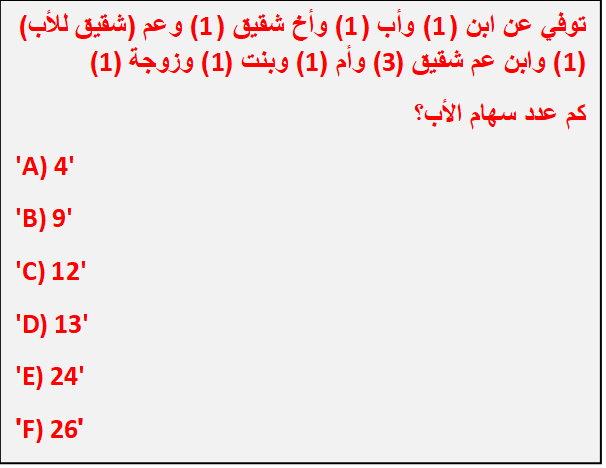}
    \caption{Example 2}
    \label{fig:exm2}
\end{figure}

\begin{figure}
    \centering
    \includegraphics[width=1\linewidth]{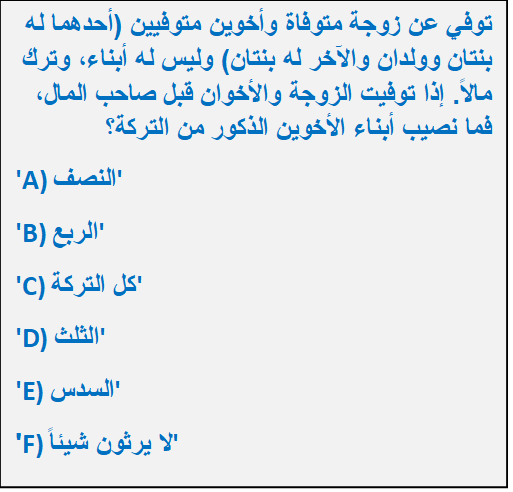}
    \caption{Example 3}
    \label{fig:exm3}
\end{figure}

The QIAS (Question-and-Answer in Islamic Studies Assessment Shared Task) 2025~\cite{qias2025,QIAS2025_2,bouchekif2024inheritance} dataset for Islamic inheritance reasoning contains multiple-choice questions (MCQs) categorized into three difficulty levels: beginner, intermediate, and advanced. The primary goal is to evaluate the reasoning abilities of LLMs within the domain of Islamic knowledge. Each question has exactly one correct answer and presents six answer choices, labeled A through F, each accompanied by a corresponding textual explanation. The dataset is annotated using one of six distinct symbols. The questions are sourced from a corpus of 32,000 fatwas from IslamWeb and have been validated by a qualified expert in Islamic inheritance law. The dataset is divided into 9,446 labeled examples for training, 1,000 labeled examples for validation, and 1,000 unlabeled examples for testing in the final test phase. Figure~\ref{fig:exm1}, Figure~\ref{fig:exm2}, and Figure~\ref{fig:exm3} show few examples of this dataset.

\subsection{Methods}

We have evaluated several methods, including base and fine-tuned LLMs, to find the best solution for Islamic inheritance reasoning. 

In the first experiment, base open-source Arabic models such as Falcon3~\cite{Falcon3,tiiuae_falcon3-7b-instruct} (``tiiuae/Falcon3-7B-Instruct''), Fanar~\cite{fanar1-9b-instruct,fanarllm2025} (``QCRI/Fanar-1-9B-Instruct''), and Allam~\cite{allam7b,bari2024allam} (``ALLaM-AI/ALLaM-7B-Instruct-preview'') were assessed. Additionally, we utilized ``Allam thinking''~\cite{almaghrabima2025allam} (``almaghrabima/ALLaM-Thinking'')~\cite{bari2024allam}, an advanced Arabic LLM, which was fine-tuned and optimized specifically for reasoning and math. It was also prompted to think step-by-step.

Additionally, proprietary models such as Gemini Flash 2.5, Gemini Pro 2.5~\cite{team2023gemini,gemini_25_thinking}, GPT-4o~\cite{GPT-4o}, and GPT o3~\cite{openai2025o3o4mini}, were evaluated for the Islamic inheritance reasoning task using the APIs of their base models. All previous LLMs were assessed in inference mode using Prompt 2 with temperature set to 0 and  top\_p set to 1.

To improve the performance of the base LLM in reasoning and increase the rate of correct answers to inheritance questions, we fine-tuned several open-source and proprietary LLMs such as GPT-4o, Gemini Flash 2.5, and Llama 4 Scout~\cite{meta2025llama4scout,meta2025llama4} (``meta-llama/Llama-4-Scout-17B-16E-Instruct''). All LLMs were fine-tuned in a supervised learning setting using the dataset mentioned above, with a training set of 7,000 examples and a validation set of 2,446 examples used during training. The results of the comparison between all LLMs, including the base and fine-tuned, are reported using the 1,000 examples in the separate validation set. 

Llama 4 was fine-tuned utilizing two prompts: Prompt 1 and Prompt 2. The fine-tuning was done using Low-Rank Adaptation (LoRA)~\cite{hu2022lora} as the Parameter-Efficient Fine-Tuning (PEFT) method. The training was carried out for seven epochs with a learning rate of 0.0002. Both the training and evaluation batch sizes were set to 1 per device, and the gradient accumulation steps were set to 1. The optimizer used was paged\_adamw\_32bit. Additionally, 10 warmup steps were used to stabilize the initial training phase. We loaded the model using 4-bit quantization  for memory efficiency. The fine-tuned models have been uploaded to Hugging Face: \url{https://huggingface.co/NYUAD-ComNets/NYUAD\_Llama4\_Inheritance\_Solver}, and \url{https://huggingface.co/NYUAD-ComNets/NYUAD\_Llama4\_Inheritance\_Solver2}.

\begin{tcolorbox}[colback=orange!5!white,colframe=orange!75!black, title={Prompt 1: Islamic inheritance reasoning}, rounded corners, boxrule=1pt, boxsep=1pt]
Answer the following question using a single word only from this list: A, B, C, D, E, F.
Final Answer: {}
\end{tcolorbox}

\begin{figure}[!b]
    \centering
    \includegraphics[width=0.9\linewidth]{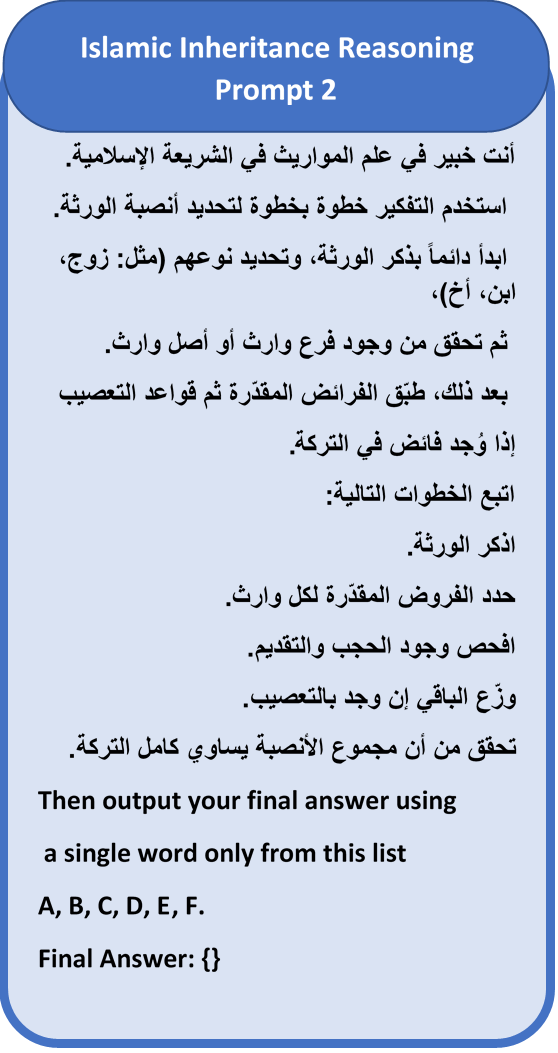}
    \label{fig:enter-label}
\end{figure}

GPT-4o was fine-tuned in two scenarios: without a system prompt and with system Prompt 2. The fine-tuning was done on the OpenAI platform  for 5 epochs, with a learning rate multiplier of 2 and automatically selected  batch size. Similarly, Gemini Flash 2 and 2.5 were fine-tuned with system Prompt 2. The fine-tuning was done on the Google AI Vertex platform. The hyper-parameters used for fine-tuning are 3 epochs and a learning\_rate\_multiplier of 5. Flash 2.5 used an adapter size of 1, while Flash 2 used an adapter size of 8 to train more parameters. In the inference phase, the threshold in safety\_settings is set to off, while thinking was enabled.

\subsection{Results and Discussion}

Table~\ref{tab:llm-accuracy} shows the accuracy of base LLMs for several open-source Arabic LLMs and proprietary state-of-the-art LLMs. Among the open models, Allam demonstrates relatively better performance (38.8\%), indicating it may be more effectively tuned for this specific task or domain. In contrast, Falcon3 and Fanar perform worse, likely due to limited domain understanding. Although ``Allam Thinking'' was optimized for reasoning and math, its accuracy declined compared to the base Allam model. This set of experiments indicates a lack of domain knowledge and reasoning in the base open-source Arabic LLMs, which 
addresses \textbf{RQ1}.

Additionally, both GPT o3 (92.3\%) and Gemini Flash 2.5 (91.5\%) demonstrate the strongest performance, highlighting their advanced capability in understanding and reasoning about Islamic inheritance, which answers \textbf{RQ2}. The small gap in accuracy may be due to the results being based on a single run. Therefore, for future work, we should run each model multiple times and compute the average and variance. This process can provide a clearer comparison between the two.

\begin{table}[h]
\centering
\begin{tabular}{lc}
\hline
\textbf{LLM} & \textbf{Accuracy (\%)} \\ \hline
Falcon3 & 24.2 \\ 
Fanar & 31.7 \\ 
Allam & 38.8 \\ 
Allam think & 29.2 \\ 
Gemini Flash 2.5 & \textbf{91.5} \\ 
GPT-4o & 70.1 \\ 
GPT O3 & \textbf{92.3} \\ \hline
\end{tabular}%
\caption{Accuracy of different Base LLMs.}
\label{tab:llm-accuracy}
\end{table}

Table~\ref{tab:llm-prompt-length} shows the prompt sensitivity of each LLM using a simple prompt such as Prompt 1 and a Chain of Thought (CoT) prompt such as Prompt 2. The GPT-4o model demonstrates a notable sensitivity to prompt design. When evaluated with Prompt 1, its accuracy is relatively low at 57.5\%. However, a shift to Prompt 2 significantly enhances the performance to 70.1\%, thereby. Allam also shows the same trend. On the other hand, Gemini Flash 2.5 exhibits high accuracy regardless of prompt content, suggesting strong internal reasoning and understanding.

\begin{table}[h]
\centering
\resizebox{\columnwidth}{!}{%
\begin{tabular}{lcc}
\hline
\textbf{LLM} & \textbf{Prompt} & \textbf{Accuracy (\%)} \\ \hline
Allam think & Prompt 1 & 28.8 \\ 
Allam think & Prompt 2  & \textbf{29.2} \\ 
Allam       & Prompt 1 & 30.4 \\ 
Allam       & Prompt 2  & \textbf{38.8} \\ 
Fanar       & Prompt 1 & 28.7 \\ 
Fanar       & Prompt 2  & \textbf{31.7} \\ 
Falcon      & Prompt 1 & \textbf{24.2} \\ 
Falcon      & Prompt 2  & 22.8 \\ 
Gemini Flash 2.5 & Prompt 1 & 90.7 \\ 
Gemini Flash 2.5 & Prompt 2  & \textbf{91.5} \\ 
GPT-4o       & Prompt 1 & 57.5 \\ 
GPT-4o       & Prompt 2  & \textbf{70.1} \\ \hline
\end{tabular}%
}
\caption{\normalsize Prompt sensitivity: Accuracy of LLMs using two different Prompts.}
\label{tab:llm-prompt-length}
\end{table}

When GPT-4o was fine-tuned for reasoning using the training set, its accuracy improved significantly, reaching over 84\% as shown in Table~\ref{tab:finetuned-llms}. On the other hand, when no system prompt is used for tuning, the accuracy reaches 84.7\%. When we added ``Prompt 2'' as a system prompt, the accuracy improved to 86.6\%. On the contrary, the performance of Gemini Flash 2.5 dropped after fine-tuning (91.5\% --> 74.6\%) on this task. This disparity in the fine-tuning results between GPT-4o and Gemini Flash 2.5 gives an answer to \textbf{RQ3}. 

The reason behind this disparity in performance after fine-tuning is that GPT-4o is a highly generalist model, not specialized for reasoning. This makes it more adaptable to niche domains like Islamic inheritance when given domain-specific data. Furthermore, GPT-4o may have moderate prior knowledge about Islamic inheritance laws, so fine-tuning filled a knowledge gap rather than conflicting with existing knowledge.

On the other hand, we observed performance degradation in fine-tuning Gemini Flash 2.5 with the adapter size set to 1 (tuning fewer parameters). To confirm our observation, we may consider fine-tuning the same model with larger adapter sizes. The reason for degradation may stem from the fact that Flash 2.5 is optimized for CoT reasoning. If the fine-tuning dataset has only final labels and lacks detailed reasoning chains, the model may lose its reasoning structure. This results in misalignment between what it's trained to do and what it's fine-tuned to.  

Fine-tuning Flash 2 with an adapter size of 8 resulted in degraded performance on this task. This may be due to the relatively limited size of the fine-tuning dataset, which was insufficient to train a larger adapter. As a result, the model likely failed to generalize well. To validate our observation, we may consider fine-tuning the same model with smaller adapter sizes or using the base model.

As shown in Table~\ref{tab:finetuned-llms}, fine-tuning GPT-4o with Prompt 2 resulted in a better performance, which contrasts with the behavior of Llama 4 Scout.

\begin{table}[h]
\centering
\begin{tabularx}{\columnwidth}{m{2.9cm}Xc}
\hline
\textbf{LLM} & \textbf{Prompt} & \textbf{Accu. (\%)} \\ \hline
Fine-tuned Llama4 Scout   & Prompt 1    & 84.3 \\ 
Fine-tuned Llama4 Scout   & Prompt 2     & 82.4 \\ 
Fine-tuned Gemini Flash 2 & Prompt 2       & 64.6 \\ 
Fine-tuned Gemini Flash 2.5 & Prompt 2     & 74.6 \\ 
Fine-tuned GPT-4o          & No system Prompt & 84.7 \\ 
Fine-tuned GPT-4o          & Prompt 2       & 86.6 \\ \hline
\end{tabularx}%
\caption{Accuracy of fine-tuned LLMs.}
\label{tab:finetuned-llms}
\end{table}

Finally, we ran an experiment to assess the best-performing base LLMs with the testing set released in the test phase and included an unlabeled set of 1000 MCQs. We ran three base models in the inference mode. The accuracies of GPT-o3, Gemini Flash 2.5, and Gemini Pro 2.5 on the testing data were 88.4\%, 88.1\%, and 87.9\%, respectively. We later applied a majority voting technique using the predictions from these three LLMs, resulting in a final accuracy of 92.7\%, which secured third place overall in the challenge.

\section*{Limitations}

The limitation of this work is that LLMs often lack comprehensive knowledge of all inheritance scenarios, which leads even the best reasoning models to occasionally select incorrect answers. Fine-tuning can help address this issue. However, a new problem arises: in the original dataset, answer choices were labeled without detailed reasoning for each option, which complicated the fine-tuning of reasoning-based models. To solve this, a second version of the dataset was released recently, including the reasoning behind each selected answer~\cite{QIAS2025_2,bouchekif2024inheritance}. This may show potential for increasing the accuracy.

\bibliography{anthology,custom}
\bibliographystyle{acl_natbib}

\end{document}